\documentclass[a4paper,twoside]{article}

\usepackage{epsfig}
\usepackage{subcaption}
\usepackage{calc}
\usepackage{amssymb}
\usepackage{amstext}
\usepackage{amsmath}
\usepackage{amsthm}
\usepackage{multicol}
\usepackage{pslatex}
\usepackage{apalike}
\usepackage[bottom]{footmisc}
\usepackage[utf8]{inputenc} % allow utf-8 input
\usepackage[T1]{fontenc}    % use 8-bit T1 fonts
\usepackage{hyperref}       % hyperlinks
\usepackage{url}            % simple URL typesetting
\usepackage{booktabs}       % professional-quality tables
\usepackage{amsfonts}       % blackboard math symbols
\usepackage{nicefrac}       % compact symbols for 1/2, etc.
\usepackage{microtype}      % microtypography

\usepackage{url}           
\usepackage{hyperref}       
\usepackage{enumitem}
\usepackage[utf8]{inputenc} 
\usepackage[frenchb, english]{babel} 
\usepackage{numprint}       
\usepackage{eurosym}        
\usepackage{appendix}       
\usepackage{xspace}         
\usepackage{xcolor}         
\usepackage{verbatim}       

\setlist[itemize]{topsep=0pt, itemsep=0pt}

\usepackage{graphicx}  
\usepackage{wrapfig}    
\usepackage{caption}    
\usepackage{subcaption} 
\usepackage{tikz}
\usetikzlibrary{positioning}
\usepackage[framemethod=TikZ]{mdframed}

\usepackage{xifthen}
\usepackage{xargs}
\usepackage{multirow} 
\usepackage{multicol}
\usepackage{booktabs} 
 
\usepackage[maxlevel=3]{csquotes} 
\usepackage{setspace}             
\usepackage{siunitx}
\usepackage{amsmath}      
\usepackage{mathrsfs}      
\usepackage{amssymb}     
\usepackage{amsfonts}      
\usepackage{amsthm}        
\usepackage{mathtools}     
\usepackage{dsfont}         
\usepackage{stmaryrd}      
\usepackage{esvect}        
\usepackage{systeme}

\usepackage{SCITEPRESS}     % Please add other packages that you may need BEFORE the SCITEPRESS.sty package.

\begin{document}

\title{Are Graph Representation Learning Methods Robust to Graph Sparsity and Asymmetric Node Information?}

 \author{\authorname{Marine Neyret\sup{*1} and Pierre Sevestre\sup{*2}}
 \affiliation{\sup{1}DataLab, Institut Louis Bachelier, Paris, France}
 \affiliation{\sup{2}DataLab, Inspection Générale et Audit, Société Générale, Paris, France}
 \affiliation{\sup{*}Both authors contributed equally.}}

\keywords{Graph Representation Learning, Transactional Graphs, Robustness, Graphs Preprocessors}

\abstract{The growing popularity of Graph Representation Learning (GRL) methods has resulted in the development of a large number of models applied to a miscellany of domains. Behind this diversity of domains, there is a strong heterogeneity of graphs, making it difficult to estimate the expected performance of a model on a new graph, especially when the graph has distinctive characteristics that have not been encountered in the benchmark yet. To address this, we have developed an experimental pipeline, to assess the impact of a given property on the models performances. In this paper, we use this pipeline to study the effect of two specificities encountered on banks transactional graphs resulting from the partial view a bank has on all the individuals and transactions carried out on the market. These specific features are graph sparsity and asymmetric node information. This study demonstrates the robustness of GRL methods to these distinctive characteristics. We believe that this work can ease the evaluation of GRL methods to specific characteristics and foster the development of such methods on transactional graphs.}

\onecolumn \maketitle \normalsize \setcounter{footnote}{0} \vfill

\section{\uppercase{Introduction}}

Graph Representation Learning (GRL) has gained increasing interest in the past few years, due to the ubiquity of its applications and the rapid development of new learning methods. First, graphs are a powerful way to represent relational data, with applications in numerous domains such as social media \cite{monti2019fake}, biology (\cite{bio1}, \cite{bio2}), knowledge graphs (\cite{kg1}, \cite{kg2}) or recommender systems \cite{Amazon}. Second, the rising interest led to a rich and evolving literature about graph learning (\cite{survey1}, \cite{survey2}), propelled by the recent development of Graph Neural Network (GNN), becoming the \textit{de facto} method for machine learning on graph, with notable contributions of Graph Convolutional Networks (GCN) \cite{gcn} and Graph Attention Networks (GAT) \cite{gat}. 

The diversity of application domains and the associated graph heteromorphism also present some drawbacks: it is difficult to foresee, from the characteristics of a graph, the expected performances of a model. In particular, a model performing well on a given graph benchmark could fail on others. Furthermore, the graph benchmark has historically been built around three citation graphs (Citeseer \cite{citeseer}, Cora \cite{cora} and Pubmed \cite{pubmed}), whose limitations, both in terms of complexity and scale, started to be highlighted (\cite{benchmarking_gnn}, \cite{salha2019simple}). Recent initiatives have been undertaken to provide more diverse and complex open-source datasets (OGB\footnote{https://ogb.stanford.edu/} and TUDataset\footnote{https://chrsmrrs.github.io/datasets/}), and to improve the quality of the benchmark with a strong emphasis on comprehensiveness and consistency (\cite{benchmarking_gnn}, \cite{benchmarking_hgnn}, \cite{benchmarking_gc}). The final objective being to accurately evaluate of models performances. Our work is in line with this philosophy, with the objective to accelerate and ease the evaluation of model performances, and more specifically to further understand how specific characteristics of a graph affect model performances. 

This can be particularly useful in the banking industry, where most of its core activities involve transactions that can be intuitively modelled using graphs. A transaction usually corresponds to a money transfer between two parties. The nodes represent the parties involved in the transactions and the edges encode the properties of the transaction between the two nodes. However, work on transactional graphs has been sporadic, mainly centered around fraud detection for financial security (\cite{transaction1}, \cite{transaction2}, \cite{transaction5}, \cite{transaction6}). The very specific nature of this application does not allow to constitute a comprehensive evaluation benchmark on transactional graphs. Furthermore, the transactional graphs obtained by the banks present distinctive characteristics which differ from common benchmark graphs. Indeed, the banks only access the transactions where their clients are involved without having access to transactions between non-clients. The resulting transactional subgraph is sparser than the complete graph of all the transactions being performed on the market. It has deteriorated local structures and a modified node distribution compared to common graphs found in the literature as shown in Appendix. A question arising is the ability of Graph Representation Learning methods to learn representations on these sparse subgraphs. Moreover, the banks hold more information about their clients than they have about the non-client counterparts. Thus, a larger range of node features are available for clients. An efficient model should be able to take advantage of this asymmetric node information. \newline

We propose an experimental pipeline to better understand how the two intrinsic characteristics (sparsity and asymmetric node information) of transactional graphs encountered within the bank affect the performance of GRL models, compared to the performance obtained on the full graphs. The objective is to analyze if node-level embeddings produced from transactional graphs encompass enough structural and semantic information about the graph to perform downstream tasks with sufficient performances. This study is performed on public benchmark graphs commonly used in the literature that are preprocessed to match the banks transactional graphs distinctive characteristics. This choice was made, in the absence of open-source datasets with the specific characteristics of transactional graphs. This allows to benefit from the knowledge acquired, especially in terms of model selection, to favor models comparison, and to isolate a specific graph property for the experiments.\newline

The contributions of this paper are as follows:
\begin{enumerate}
    \item \textbf{Challenge some Graph Representation Learning methods on transactional graph distinctive characteristics.} We study the effect of the two banks transactional graphs distinctive characteristics that are graph sparsity and asymmetric node information on the performance of state-of-the-art models, through the transformation of commonly used graphs in the literature to match these characteristics.
    \item \textbf{Design of a robust experimental pipeline.} We develop a standardized pipeline to easily experiment various models and graph configurations. \newline
\end{enumerate}
%\footnote{Implementation will be available publicly by the time of the conference.}
The rest of the paper is organized as follows. Section \ref{sec:deux} introduces definitions about graph theory. Section \ref{sec:trois} defines the preprocessors designed to transform commonly used graphs and gives some theory about the surveyed Graph Representation Learning methods. Our experimental results are detailed in Section \ref{sec:quatre}.

\section{\uppercase{Definitions and preliminaries}}
\label{sec:deux}

In this paper, we consider a graph $\mathcal{G} = (\mathcal{V}, \mathcal{E})$, with $\mathcal{V} = \{v_1, ..., v_n \}$ its set of $n$ vertices (or nodes), and $\mathcal{E} \subset \mathcal{V} \times \mathcal{V}$, its set of $m$ edges. 
When node attributes are available, the graph can be associated with a node features matrix $X \in \mathbb{R}^{n \times f}$, with $f$ being the node features dimension. We focus in the following on undirected and unweighted graphs, containing neither self-loops nor multiple edges.\newline

We denote by 
\begin{equation}
    A = (a_{ij})_{i, j=1, ..., n}
\end{equation}
 the \textit{adjacency matrix} representing the graph structure, with $a_{ij} = 1$ if an edge exists between nodes $v_i$ and $v_j$, and $a_{ij} = 0$ otherwise.

 \begin{equation}
    \tilde{A} = A + I_n
\end{equation}
corresponds to the adjacency matrix of the graph with all self-loops included.

We further define $d_i$, the degree of node $v_i \in \mathcal{V}$, as the sum of neighboring weights, i.e. the number of neighbors in the case of an unweighted graph, $d_i = \sum_{j=1}^{n} a_{ij}$. The \textit{degree matrix} is then defined as the diagonal matrix containing all node degrees, 
\begin{equation}
    D = diag(d_1, ..., d_n).
\end{equation} 
The degree matrix of $\tilde{A}$ is denoted by $\tilde{D}$.

From the two matrices introduced above, we derive the \textit{unnormalized graph Laplacian matrix}, defined as: 
\begin{equation}
    L = D - A.
\end{equation}

Table \ref{tab:notations} summarizes the main notations used throughout the paper.

\begin{table*}[h!]
    \caption{Summary of commonly used notations.\\}
    \label{tab:notations} 
    \centering
    \begin{tabular}{@{}ll@{}}
    \toprule
    \textbf{Notation} &  \textbf{Description} \\ \midrule
    $\mathcal{G}$, $\mathcal{V}$, $\mathcal{E}$ &  Graph, set of nodes, set of edges\\ 
    $n$, $m$ & Number of nodes and edges\\
    $v_i$, $\mathcal{N}_i$& Node $i$ and its first-order neighborhood\\
    $A$, $D$, $L$ & Adjacency matrix, degree matrix, Laplacian\\ 
    $d_i, d$ & Degree of node $v_i$, average degree\\ 
    $X$, $f$ & Node features matrix, node features dimension\\ 
    $\tilde{A}$, $\tilde{D}$ & Adjacency and degree matrices of the graph with all self-loops \\
    a, $\mathbf{a}$, A & Scalar, vector, matrix \\
    $\mathbb{R}^d$ & Euclidean space, d-dimensional\\
    $A^{\top}$ & Transposed of $A$\\
    $\bigparallel$ & Concatenation operation\\
    $H^l$, $H^l_i$ & Hidden representation matrix and hidden representation of node $v_i$ at layer $l$\\
    $f_l$ & Hidden representation dimension at layer $l$
    \end{tabular}
\end{table*}

\section{\uppercase{Proposed benchmark method}}
\label{sec:trois}

The proposed experimental pipeline consists in two parts. The first part, which we refer to as the preprocessor, corresponds to the transformations applied to the graphs to match certain distinctive characteristics. Further details can be found in Section \ref{sec:troisun}. Then follows the model training and evaluation part. Assessing the quality of the learned node embeddings is usually reduced to the ability to perform a given downstream task. In this study, we chose to focus on the node classification and link prediction tasks, as described in Section \ref{sec:troisdeux}.

\subsection{Preprocessors}
\label{sec:troisun}

\begin{figure*}[h!]
    \centering
    \includegraphics[width=0.8\linewidth]{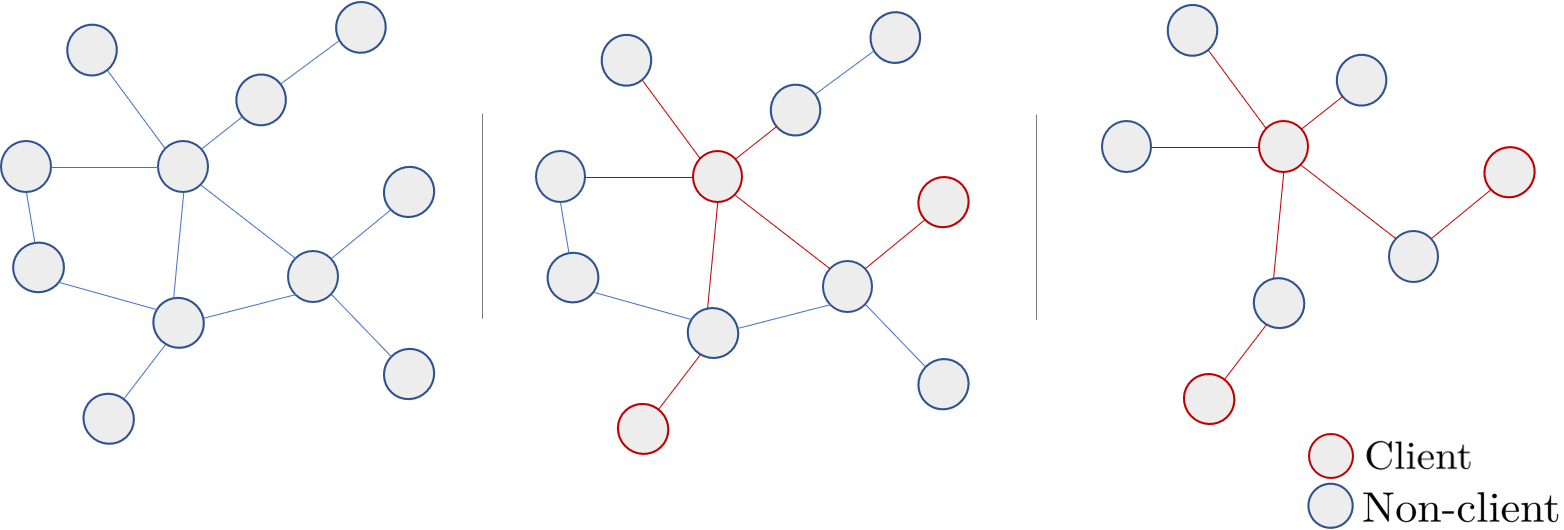}
    \caption{Illustration of the sampling procedure. Client nodes are first sampled according to $r_{sampling}$, then, all edges not connected to these nodes are removed.}
    \label{fig:sampler}
\end{figure*}

As mentioned previously, this study intends to evaluate the impact of the banks transactional graph singularities, that are partial graph information, or sparsity, and asymmetric node information.
To achieve this, two graph preprocessors are designed to transform original graphs from the literature to match the banks transactional graphs distinctive features:

\begin{itemize}
    \item The first processor is targeting solely the sparsity aspect. The procedure, described on Figure \ref{fig:sampler}, consists in creating a subgraph of the original graph, by first sampling nodes that will act as client nodes and then removing any edge not connected to this subset of client nodes. We will refer to this processor as the \textit{sampler}, and to the corresponding portion of client nodes as $r_{sampling}$. 
    \item The second processor further targets the asymmetric node information. Here, the \textit{sampler} is first applied. Node features of sampled clients nodes are left unchanged and we deteriorate a portion of the node features of the remaining set of nodes. In the reported results, the deterioration method simply consists in the deletion of a portion of node features. This processor will be referred to as \textit{features\_sampler}, and we will use $r_{sampling}$ to invoke the portion of sampled client nodes and $r_{nf\_sampling}$ for the portion of deteriorated node features. \newline
\end{itemize}

In the experiments, we will use the term \textit{baseline} when no processor is applied, i.e. when the entire graph is provided to the model. A decreasing $r_{sampling}$ is used with values in ($0.5$, $0.1$, $0.01$, $5e-3$, $1e-3$), provided that the number of remaining nodes from the original graph is sufficient. $r_{nf\_sampling}$ is chosen among ($0.1$, $0.5$, $0.75$), in order to monitor the impact of the two distinctive characteristics.

\subsection{Models Surveyed}
\label{sec:troisdeux}

Robustness to these graph distinctive characteristics of both unsupervised, random-walk based and supervised graph neural networks based methods are assessed. To each method corresponds a dedicated evaluation pipeline based on downstream tasks detailed below. The evaluation of the models focuses on nodes-centric tasks, more precisely node classification and link prediction. Node classification stands for supervised learning of node labels. Link prediction refers to the task of predicting the existence of an edge between source and destination nodes.  

\subsubsection{Unsupervised Models}

\textbf{DeepWalk} operates on the graph structure only, without taking nodes features into account. Starting from the observation that the distribution of nodes in a corpus of small random walks follows the same power-law distribution as the one observed for words in a text corpus, \cite{deepwalk} introduced DeepWalk and used the progress in word representation \cite{word2vec} to create node representation. 
The algorithm first transforms the graph sequences of nodes using truncated, unbiased random walks (corresponding to the text corpus for a word representation). Then, this set of walks is fed to Skipgram \cite{word2vec}, that learns to predict context nodes in random walks for each target node. The resulting representation creates close embedding for nodes co-occuring in random walks, preserving high-order proximity.\newline

\noindent \textbf{node2vec} \cite{node2vec} is derived from DeepWalk. The difference lies on the random walks generation, through the use of biased random walks to give more flexibility to the exploration. Specifically, two parameters allow the exploration to favor either Breadth-first or Depth-first strategy. These walks create a new definition of neighborhood, and the resulting embedding no longer depends only on the community, but on the local structure around the node as well.\newline

\noindent \textbf{Training and evaluation processes.} Once embeddings are computed from these unsupervised models, the pipelines are split according to the evaluation task. The unsupervised embeddings are used for two distinct downstream tasks: node classification and link prediction. For node classification, a logistic regression is trained using node embeddings as input, to predict some node labels. For link prediction, nodes embeddings of two nodes are concatenated and a logistic regression is trained to predict the likelihood of these two nodes being connected. Existing and non-existing edges are sampled evenly.

\subsubsection{Supervised Models}
\textbf{Graph Convolutional Networks (GCN)} iteratively apply spatial filter on one node's direct neighbors information to produce its embedding. This model takes into account both graph structure and node features, and it assumes that graph edges encode nodes similarity.

The propagation rule from \cite{gcn} focuses on first order approximation and is defined by:
\begin{equation}
H^{l+1} = \sigma \left(\widetilde{D}^{-\frac{1}{2}} \widetilde{A} \widetilde{D}^{-\frac{1}{2}} H^l \Theta^l \right),
\end{equation}

where $\sigma(.)$ is a non-linear transformation, $\Theta^l$ a trainable filter matrix, of shape $f_l \times f_{l+1}$ and $H^l \in \mathbb{R}^{n \times f_l}$ the hidden state at layer l ($f_{l}$ being the embedding dimension at layer l and $H^0$ being initialized to X).

At each layer, the embedding of one node corresponds to a combination of surrounding nodes embeddings with pre-determined weights depending on nodes degrees.\newline

\noindent \textbf{GraphSage} \cite{graphsage} learns a representation of one node based on an aggregation of a fixed $K$-hop neighborhood information. For each node, the embedding is also initialized to the node features matrix and, then, the model iteratively creates an aggregated embedding using $k<K+1$ depth neighbors embeddings that is concatenated with the embeddings computed from smaller neighborhoods. All the neighboring nodes share the same weight in the embedding.

For a node $v_i$, using the same notations as for GCN, the embedding at layer $l+1$ is given by:
\begin{align}
    a^{l+1}_{\mathcal{N}_i} &= \texttt{agg}\left(\{H^{l}_{u}, \forall u \in \mathcal{N}_i\}\right),\\
    H^{l+1}_{i} &= \sigma \left( \Theta^l * \left(H^{l}_{i} \bigparallel a^l_{\mathcal{N}_i}\right)\right).
\end{align}

The authors of the paper proposed to use either a mean, a LSTM or a pooling aggregator ($\texttt{agg}$ in previous equation). We focused on the mean aggregation function only.\newline

\noindent \textbf{Graph Attention Networks (GAT)} \cite{gat} integrates attention related strategy in their layers to evaluate connectivity strength between nodes. The idea behind GAT is to employ a self-attention mechanism to compute the representation of a node by attending over all its neighbors, learning to pick the most relevant ones.

The output at layer $l+1$ is a weighted linear combination of the previous hidden state representations:
\begin{equation}
    H^{l+1}_i = \sigma \left( \sum_{j \in \mathcal{N}_i} \alpha^{l}_{ij} \Theta^l H^l_i\right),
\end{equation}

where $\alpha^{l}_{ij}$ is the normalized attention coefficient between node $v_i$ and $v_j$ at layer $l$:
\vspace{0.2cm}
\begin{align}
    \alpha^{l}_{ij} &= \texttt{softmax}_{\mathcal{N}_i}\left(e^{l}_{ij}\right),\\
    e^{l}_{ij} &= \texttt{LeakyReLU} \left(\mathbf{a}^{\top}\left[\Theta^{l} H_i \bigparallel \Theta^{l} H_j \right]\right).
\end{align}

\noindent \textbf{Training and evaluation processes.} For node classification, no stage is added as the models are directly trained to classify the nodes according to their labels. For link prediction, a 3-layer perceptron is stacked to the models and trained using contrastive loss, with negative graphs generated at every epoch and the models learning to discriminate existing edges from the graph and non-existing edges from the negative graph. 

\subsection{Datasets}
\label{sec:data}

For this study, GRL robustness assessment is performed on the four datasets described below. This choice was motivated by the need to use graphs large enough to be able to sample nodes in substantial proportion preserving decent scale and complexity for the resulting graph. The selected datasets are: 

\begin{itemize}
\item \textbf{Pubmed} \cite{pubmed}, an undirected citation network of diabetes-related publications, with nodes corresponding to research papers authors and edges to citation relationship. The 3 available labels are derived from the type of diabetes addressed in the publication, and node features are sparse bag-of-words of dimension 500.
\item \textbf{CoAuthor CS}, a citation network within the field of computer science, created using \textit{\href{https://www.microsoft.com/en-us/research/project/microsoft-academic-graph/}{Microsoft Academic Graph} from the \href{https://www.kdd.org/kdd-cup/view/kdd-cup-2016}{KDD Cup 2016}}. Nodes are also authors, and edges exist if authors co-authored a paper. The corresponding node label is assigned based on the most active field of research of the author for a total of 15 classes. Each node is associated with a 6,805 dimensional features vector representing papers keywords.
\item \textbf{Amazon Computer} \cite{Amazon}, a recommendation network extracted from Amazon co-purchase graph. Nodes are products, and edges indicate whether these products are frequently bought together. The 5 classes are derived from the product categories, and node features from bag-of-words encoded product reviews of dimension 767. 
\item \textbf{Reddit} \cite{graphsage}, a social network created from posts made in September 2014 on the online service. Nodes correspond to posts, each belonging to a community, or \textit{subreddit}, assigned as label. The network has 50 labels. Edges between posts are created if a user of the platform commented both posts. Nodes are described by 602 features containing a concatenation of an average embedding of dimension 300 of the post title, an average embedding of dimension 300 of the post's comments, the post's score and the number of post's comments.
\end{itemize}

\section{\uppercase{Experiment pipeline results}}
\label{sec:quatre}

%The figure is positionned here in the code to force its position on page 6 in the pdf
\begin{figure*}[h!]
  \captionsetup[subfigure]{justification=raggedright, singlelinecheck=false}
\begin{subfigure}{.5\textwidth}
  \centering
  \caption{}
  \begin{tikzpicture}
    \node (img)  {\includegraphics[scale=0.49]{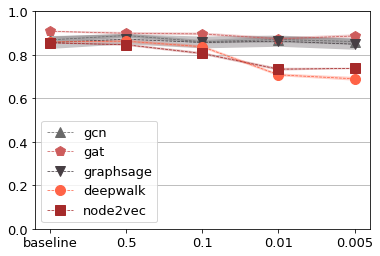}};
    \node[below=of img, node distance=0cm, xshift=0.4cm, yshift=1.2cm] {$r_{sampling}$};
    \node[left=of img, node distance=0cm, rotate=90, anchor=center, yshift=-0.9cm] {Accuracy};
   \end{tikzpicture}
  \label{fig:nc_sampler}
\end{subfigure}%
\hfill
\begin{subfigure}{.5\textwidth}
  \centering
  \caption{}
  \begin{tikzpicture}
    \node (img)  {\includegraphics[scale=0.49]{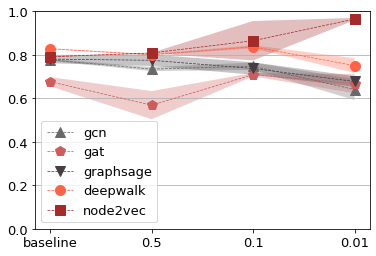}};
    \node[below=of img, node distance=0cm, xshift=0.4cm, yshift=1.2cm] {$r_{sampling}$};
    \node[left=of img, node distance=0cm, rotate=90, anchor=center,yshift=-0.9cm] {AUC};
   \end{tikzpicture}
  \label{fig:lp_sampler}
\end{subfigure}
\caption{(a) Node classification accuracy of surveyed models using the \textit{sampler} preprocessor on Amazon Computer. (b) Link prediction AUC of surveyed models using the \textit{sampler} preprocessor on Pubmed. Both for (a) and (b), results are averaged over 100 independent runs (10 trainings on 10 generated graphs) and reported for the test set. Line thickness accounts for the standard deviation.}
\label{fig:fig}
\end{figure*}
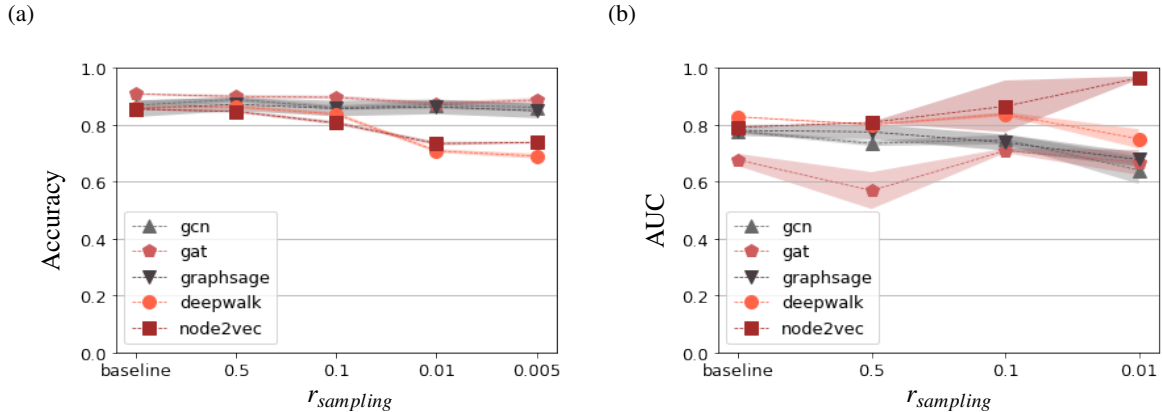

\subsection{Experimental Setting}

We implemented our own version of DeepWalk. Default parameters defined in the original paper are applied to report results: 10 walks of length 80 are generated from each node and the resulting node embedding is of dimension 128.\newline

The nodevectors\footnote{https://github.com/VHRanger/nodevectors} fast and scalable implementation of node2vec is used. Walks length, walks number and embedding dimension are set to the same values as for DeepWalk. According to hyperparameters sensitivity analysis presented in \cite{survey1}, we set $(p, q) = (1, 2)$ for link prediction and $(p, q) = (1, 0.5)$ for node classification. \newline

The three graph neural network models under review are implemented using Deep Graph Library (DGL) \cite{dgl}. For each dataset, the models parameters are gridsearched on the baseline graph and then used on all sampling experiments. Although it would have been more accurate to search for optimal parameters on each graph configuration, the associated computation cost was high, motivating the use of this approach. \newline

Experiments are conducted on \textit{NVIDIA Tesla V100-PCIE-16GB}. 

\subsection{Impact of Graph Sparsity on Models Performances}

\textbf{Node classification.}
Although models are built upon various approaches, the \textit{sampler} preprocessor has limited, consistent effect over all models and datasets. If we ignore a few exceptions encountered for Pubmed that has a small number of nodes, the drop in performance when sampling stays below $11\%$ compared to the unsampled baseline for supervised graph neural networks. This value drops to $4\%$ when restricting $r_{sampling}$ to values bigger than $0.01$. The change is even less significant on CoAuthor CS and Amazon Computer. Random walks based unsupervised methods tend to be more impacted by the \textit{sampling} preprocessing than graph neural networks. The first models cannot use node features to compensate for the loss of information from the sampling, nor benefit from the supervised signal of the remaining models. This can be seen in Figure \ref{fig:nc_sampler}.\newline

\noindent \textbf{Link prediction.}
Experiments on link prediction provide less consistent results across models and datasets: the performance when sampling can either stay stable, increase or decrease, with higher variability than on node classification. Convergences issues are faced for some experiments with GAT contributing to results discrepancy (see Figure \ref{fig:lp_sampler}). The unbalanced effect between random walk based models and supervised graph neural networks no longer stands out for link prediction.

Setting aside the evaluation task, when the number of nodes in the whole graph is limited, as for Pubmed, the models behavior becomes unstable when sampling ratio is decreasing: the performance can increase for small ratios as seen in Figure \ref{fig:lp_sampler}, and the drop in performance can soar (up to$80\%$ below baseline) in node classification.

\subsection{Impact of Asymmetric Node Information on Models Performances}

As DeepWalk and node2vec do not introduce node features into embeddings computation. Only the robustness evaluation to asymmetric node information of supervised graph neural networks is performed.\newline

\begin{table*}[h!]
    \caption{Results of surveyed models in terms of node classification accuracies using the \textit{features\_sampler} preprocessor on Amazon Computer. Results are averaged over 100 independent runs (10 trainings on 10 generated graphs) and standard deviations are reported.}
    \label{blur_sampler_nc}
    \centering
    \begin{tabular}{lllll}
      \toprule
      \multicolumn{2}{c}{Amazon Computer} & \multirow{2}{*}{GCN} & \multirow{2}{*}{GAT} & \multirow{2}{*}{GraphSage}\\
      \cmidrule(r){1-2}
      $r_{nf\_sampling}$     & $r_{sampling}$     & & & \\
      \midrule
      \multicolumn{2}{c}{\textit{baseline}} & 0.869 $ \pm 0.015$ & 0.908 $ \pm 0.003$ & 0.857 $ \pm 0.030$ \\
      \midrule
      \multirow{4}{*}{0.1}  & 0.5   & 0.882 $ \pm 0.013$ & \textbf{0.910} $ \pm 0.004$ & \textbf{0.871} $ \pm 0.026$ \\
                            & 0.1   & 0.886 $ \pm 0.018$ & 0.895 $ \pm 0.005$ & 0.861 $ \pm 0.022$ \\
                            & 0.01  & 0.875 $ \pm 0.016$ & 0.874 $ \pm 0.004$ & 0.848 $ \pm 0.030$\\
      \midrule
      \multirow{4}{*}{0.5}  & 0.5   & \textbf{0.896} $ \pm 0.004$ & 0.900 $ \pm 0.004$ & 0.852 $ \pm 0.023$\\
                            & 0.1   & 0.884 $ \pm 0.004$ & 0.889 $ \pm 0.008$ & 0.855 $ \pm 0.017$ \\
                            & 0.01  & 0.879 $ \pm 0.009$ & 0.883 $ \pm 0.006$ &  0.833 $ \pm 0.020$\\
      \midrule
      \multirow{4}{*}{0.75} & 0.5   & 0.878 $ \pm 0.002$ & 0.888 $ \pm 0.009$ & 0.814 $ \pm 0.022$ \\
                            & 0.1   & 0.857 $ \pm 0.008$ & 0.866 $ \pm 0.006$ & 0.792 $ \pm 0.032$\\
                            & 0.01  & 0.850 $ \pm 0.004$ & 0.841 $ \pm 0.009$ & 0.774 $ \pm 0.025$ \\
      \bottomrule
    \end{tabular}
\end{table*}

\begin{table*}[h!]
  \caption{Results of surveyed models in terms of link prediction AUC using the \textit{features\_sampler} preprocessor on CoAuthor CS. Results are averaged over 100 independent runs (10 trainings on 10 generated graphs) and standard deviations are reported.}
  \label{blur_sampler_lp}
  \centering
  \begin{tabular}{lllll}
    \toprule
    \multicolumn{2}{c}{CoAuthor CS} & \multirow{2}{*}{GCN} & \multirow{2}{*}{GAT} & \multirow{2}{*}{GraphSage}\\
    \cmidrule(r){1-2}
    $r_{nf\_sampling}$     & $r_{sampling}$     & & & \\
    \midrule
    \multicolumn{2}{c}{\textit{baseline}} & \textbf{0.792} $\pm 0.018$ & \textbf{0.851} $\pm 0.074$ & \textbf{0.933} $\pm 0.009$ \\
    \midrule
    \multirow{3}{*}{0.1}  & 0.5   & 0.713 $\pm 0.029$ & 0.758 $\pm 0.119$ & 0.922 $\pm 0.008$ \\
                          & 0.1   & 0.731 $\pm 0.037$ & 0.738 $\pm 0.106$ & 0.907 $\pm 0.008$ \\
                          & 0.01  & 0.699 $\pm 0.022$ & 0.778 $\pm 0.078$ & 0.891 $\pm 0.009$  \\
    \midrule
    \multirow{3}{*}{0.5}  & 0.5   & 0.734 $\pm 0.041$ & 0.670 $\pm 0.144$ & 0.921 $\pm 0.010$ \\
                          & 0.1   & 0.709 $\pm 0.022$ & 0.672 $\pm 0.138$ & 0.896 $\pm 0.006$ \\
                          & 0.01  & 0.699 $\pm 0.023$ & 0.610 $\pm 0.116$ & 0.883 $\pm 0.011$ \\
    \midrule
    \multirow{3}{*}{0.75} & 0.5   & 0.723 $\pm 0.024$ & 0.635 $\pm 0.138$ & 0.901 $\pm 0.011$ \\
                          & 0.1   & 0.696 $\pm 0.020$ & 0.650 $\pm 0.128$ & 0.877 $\pm 0.016$ \\
                          & 0.01  & 0.684 $\pm 0.007$ & 0.644 $\pm 0.099$ & 0.860 $\pm 0.018$ \\
    \bottomrule
  \end{tabular}
\end{table*}

\noindent \textbf{Node classification.}
Similar conclusions to these obtained for the \textit{sampler} preprocessor can be drawn about the impact of the \textit{features\_sampler} (see Table \ref{blur_sampler_nc}). Out of most experiments, with constant $r_{nf\_sampling}$ (\textit{resp.} constant $r_{sampling}$), accuracies decrease as $r_{sampling}$ decreases (\textit{resp.} $r_{nf\_sampling}$ increases) and the performance drop lies into the $0 - 13\%$ range. The least variable results are obtained on CoAuthor CS dataset as the change is $1.5\%$ at most. Both preprocessors have similar and weak effect on models performances. Regarding Reddit and Amazon Computer, the \textit{features\_sampler} preprocessor has almost always 1.5 times more impact than the \textit{sampler}. Finally, GCN is less influenced by nodes features deletion than GAT and GraphSage. Results on Pubmed also suffer from inconsistencies with low values of $r_{sampling}$. \newline

\noindent \textbf{Link prediction.}
As for the \textit{sampler} preprocessor, no matter what $r_{nf\_sampling}$ is used, no global trend is identified as $r_{sampling}$ decreases. Results are summarized in Table \ref{blur_sampler_lp}. However, with constant $r_{sampling}$, the impact of $r_{nf\_sampling}$ is limited over all datasets and models. Setting aside experiments with convergence issues, the drop in performance stays below $9\%$ compared to the sampled experiment with the corresponding $r_{sampling}$ and $15\%$ compared to the unsampled baseline.

\section{\uppercase{Conclusion}}

We developed an experimental pipeline to ease the evaluation of Graph Representation methods against distinctive properties, and used this pipeline for the particular case of transactional graphs encountered in the banking industry. This leads to the experimentation around the two properties that are graph sparsity and asymmetric node information. We demonstrated that Graph Representation Learning methods are robust to graph sparsity and asymmetric node information when performing node classification. For link prediction, although our study does not allow to draw a unified conclusion due to unstable training encountered in many graph configurations, we did not observe significant performance decrease even in harsh preprocessing configurations.

\section*{\uppercase{Acknowledgements}}
This research was conducted within the "Data Science for Banking Audit" research program, under the aegis of the Europlace Institute of Finance, a joint program with the General Inspection at Société Générale.

\bibliographystyle{apalike}
{\small
\bibliography{main}}

\section*{APPENDIX}

\begin{table*}[t!]
    \caption{Detailed datasets statistics on full graphs and sampled versions. Real transactional graph provided as example.}
    \label{tab:stats_sampling}
    \centering
    \begin{tabular}{@{}lrrrrrrrr@{}}
        \toprule
    \textbf{Name} & $r_{sampling}$ & $n$ & $m$ & $d$ & $p$ & $\overline{t}$  & $\gamma$ & $H_{er}$ \\ 
    \midrule
    \textbf{Transactional graph} &  &  140,674 & 190,371 & 2.71 & 1.9e-5 & 0.24 & 5.29 & 0.70 \\
    \cmidrule{1-9}
     \multirow{4}{*}{Pubmed} & 1 & 19,717 & 44,327 & 4.50 & 2.3e-4  & 1.90 & 2.18 & 0.93 \\
     & 0.5 &  16,884 & 33,166 & 3.93 & 2.3e-4 & 1.07 & 2.28 & 0.93 \\
     & 0.1 &  7,216 & 8,840 & 2.45 & 3.4e-4 & 0.15 & 3.17 & 0.93  \\
     & 0.01 &  1,083 & 988 & 1.82 & 1.7e-3& 0.05 & 5.34 & 0.91 \\
     \cmidrule{1-9}
     \multirow{4}{*}{CS} & 1 &  18,333 & 81,894 & 8.93 & 4.9e-4 & 14.04 & 1.55 & 0.96 \\
     & 0.5 &  17,638 & 61,828 & 7.01 & 4.0e-4 & 7.41 & 1.65 & 0.96 \\
     & 0.1 &  10,269 & 16,154 & 3.15 & 3.1e-4 & 0.77 & 2.48 & 0.94 \\
     & 0.01 &  1,656 & 1,604 & 1.94 & 1.2e-3& 0.06 & 5.03 & 0.91 \\
     \cmidrule{1-9}
     \multirow{4}{*}{Amazon} & 1 &  13,471 & 245,861 & 36.50 & 2.7e-3 & 340.17 & 1.33 & 0.93 \\
     & 0.5 &  13,288 & 182,199 & 27.42 & 2.1e-3 & 164.72 & 1.38 & 0.93 \\
     & 0.1 &  10,856 & 4,685 & 8.63 & 8.0e-4 & 10.64 & 1.75 & 0.89 \\
     & 0.01 &  2,927 & 3,982 & 2.72 & 9.3e-4 & 0.20 & 4.00 & 0.84 \\
     \cmidrule{1-9}
     \multirow{4}{*}{Reddit} & 1 &  232,965 & 57,307,946 & 491.99 & 2.1e-3 &  1.08e5 & 1.19 & 0.94 \\
     & 0.5 &  232,036 & 43,064,864 & 371.19 & 1.6e-3 & 5.42e4 & 1.20 & 0.94 \\
     & 0.1 &  222,650 & 10,960,320 & 98.45 & 4.4e-4 & 3.24e3 & 1.30 & 0.90 \\
     & 0.01 &  172,492 & 1,174,440 & 13.62 & 7.9e-5 & 48.39 & 1.70 & 0.82 \\
    \bottomrule
    \end{tabular}
    
    \end{table*}

In Table \ref{tab:stats_sampling}, detailed statistics about all datasets under study are provided, along with their sampled version according to the sampling procedure defined in Section \ref{sec:troisun}. \newline

$n$, $m$ and $d$ are defined in Table \ref{tab:notations}. \newline

The fill, referred to as $p$, denotes the probability that an edge is present between two randomly chosen nodes. For an undirected graph without loops, we have:
\begin{equation}
    p = \frac{2m}{n(n-1)}.
\end{equation}

\begin{figure}[h!]
    \centering
    \includegraphics[width=0.1\textwidth]{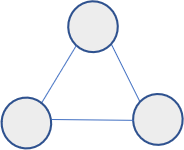}
    \caption{Illustration of a triangle pattern.}
    \label{fig:triangle}
\end{figure}

$\overline{t}$ corresponds to the number of triangles (see Figure \ref{fig:triangle}), or 3-cycle, normalized by the number of nodes, $n$.

The two other statistics are derived solely from the distribution of degree values $(d_i)_{i=1, ..., n}$, without considering the graph structure. It can be shown that most often, the distribution of degree of any graph follows a power law distribution, meaning that the number of nodes with degree $n$ is proportional to $n^{-\gamma}$, with $\gamma > 1$. The power law exponent \textit{gamma} is the most notable statistics. The closer the value is to one, the more skewed towards higher degree the distribution is, thus the more nodes with high degree the graph possesses. On the contrary, higher values indicates that the number of neighbors falls rapidly in the graph.   \newline
This $\gamma$ value is often estimated using the following robust estimator \cite{Newman_2005}:

\begin{equation}
\gamma = 1 + n \left( \sum_{v_i \in \mathcal{V}} \ln{\frac{d_i}{d_{min}}} \right)^{-1},
\end{equation}

where $d_{min} = \underset{v_i \in \mathcal{V}}{\min d_i}$

Another metric derived from the degree distribution is the relative edge distribution entropy, measuring the equality of the degree distribution. This statistics approaches zero when all edges are attached to a single node, and equals one when all nodes have equal degree. It is defined as \cite{Kungis_2012}:

\begin{equation}
H_{er} = \frac{1}{\ln n} \sum_{v_i \in \mathcal{V}} -\frac{d_i}{2m} \ \ln{\frac{d_i}{2m}}.
\end{equation}

\end{document}